\documentclass{mva_style}
\usepackage{graphicx}
\usepackage{color}
\usepackage{svg}
\usepackage{url}
\usepackage{algorithm}
\usepackage{algpseudocode}

\finalcopy 

\begin{document}
\title{
Real-Time LiDAR Point Cloud Densification \\for Low-Latency Spatial Data Transmission
}

\author{
  Kazuhiko Murasaki, Shunsuke Konagai, Masakatsu Aoki, Taiga Yoshida, Ryuichi Tanida\\
  NTT Human Informatics Laboratories
}

\maketitle

\section*{\centering Abstract}
\textit{
To realize low-latency spatial transmission system for immersive telepresence,
there are two major problems:
capturing dynamic 3D scene densely and processing them in real time.
LiDAR sensors capture 3D in real time, but produce sparce point clouds.
Therefore, this paper presents a high-speed LiDAR point cloud densification method 
to generate dense 3D scene with minimal latency, 
addressing the need for on-the-fly depth completion while maintaining real-time performance. 
Our approach combines multiple LiDAR inputs with high-resolution color images 
and applies a joint bilateral filtering strategy implemented through 
a convolutional neural network architecture. 
Experiments demonstrate that the proposed method produces dense depth maps 
at full HD resolution in real time (30 fps),
which is over 15x faster than a recent training-based depth completion approach \cite{BPNet}.
The resulting dense point clouds exhibit accurate geometry 
without multi-view inconsistencies or ghosting artifacts. 
}

\section{Introduction}
\vspace{-2mm}

In recent years, many projects have explored capturing the entirety of live events (e.g. music concerts or sports) 
in both time and space and transmitting that data to remote locations to provide viewers 
with a sense of immersion as if they were physically present \cite{holoportation,dou2016fusion4d}. 
We refer to a system that realizes such a service as a spatial transmission system. 
A spatial transmission system needs to capture a 3D space in real time, 
transmit it with low latency, and present it to users as a 3D representation. 
Among these requirements, the particularly challenging issue is 
the real-time capture of a dense 3D space.

Volumetric video capture \cite{volumetric} can reconstruct 3D information from numerous camera feeds and enable free-viewpoint rendering, 
but it requires several seconds of processing and thus cannot achieve the low latency 
needed for interactive experiences.
Moreover, they focus on only human's behavior, not space.
Alternatively, using 3D sensors like LiDAR allows real-time 3D capture; 
however, LiDAR data is much sparser than camera images. 
Achieving an immersive level of detail with LiDAR alone would require 
a very large number of sensors, which is impractical.

In this work, we aim to realize a spatial transmission system that maintains real-time 
performance while achieving a dense 3D capture sufficient for immersion. 
To this end, we propose a method to rapidly densify 3D point clouds acquired by LiDAR, 
increasing their density without sacrificing the real-time requirement.

\section{System Overview}
\vspace{-2mm}

We first outline the system used in our experiments. 
To capture the space in real time, 
multiple cameras and LiDAR units are arranged around the target area, 
and their relative positions are calibrated in advance. 
Each camera provides 1920×1080 RGB images at 30 fps. 
At each viewpoint, we use three LiDAR sensors operating at 10 fps, 
staggered in time, 
to achieve a combined effective depth frame rate of 30 fps. 
In other words, by interleaving three LiDAR scans, we obtain a pseudo-30 fps colored depth stream for that viewpoint. 
Each viewpoint has a dedicated PC that processes its camera and LiDAR data. 
The LiDAR point clouds from the three scans are merged and projected into the camera view to form a single depth image. 
This depth image synthesis is performed in parallel per viewpoint. 

To integrate data from multiple viewpoints, we simply overlay the colored point clouds 
from each viewpoint into a common 3D coordinate system. 
Notably, there is no information exchange between different viewpoint PCs; 
this design prioritizes low latency by avoiding cross-network communication. 
The fused multi-view point cloud represents the entire captured space in real time. 


\section{Related Work}
\vspace{-2mm}

Reconstructing a dense depth map from sparse depth data with 
the help of a high-resolution RGB image is called guided depth completion, and has been extensively studied
\cite{DepthCompSurvey2022,RGBGuidedSurvey2024}. 
For example, DAGF \cite{DAGF} employs a guided filter-based CNN model that uses the RGB image as 
guidance to upsample a low-resolution depth image. 
This approach works well when the sparse depth samples lie on a regular grid.
However, in our system the LiDAR’s viewpoint is different from the camera’s, 
so the LiDAR depth points appear at random positions 
when projected onto the camera image. 
Methods that assume grid-structured depth input, like DAGF, 
cannot be directly applied in this scenario.

Recently, many approaches try to densify 
randomly sparse depth with RGB image as guidance \cite{RGBGuidedSurvey2024},
and some of them have achieved real-time processing \cite{BPNet, bai2020depthnet, long2024sparsedc}.
BP-Net \cite{BPNet} builds a CNN model that propagates 
the sparse depth information to neighboring pixels, 
guided by the RGB image, to fill in missing depths. 
It achieves high accuracy and is faster than many prior methods. 
However, BP-Net still requires about 20 ms to process a 320×240 image, 
which is insufficient for our goal of 30 fps at Full HD resolution.
Moreover, most of existing methods rely on
supervised learning with high-density depth data for training. 
In practice, obtaining such ground-truth dense depth for training is difficult, 
and methods trained on one domain (e.g. indoor datasets like NYU Depth \cite{NYUDepthV2} ) may not 
generalize robustly to different environments. 
This need for extensive training data and the potential domain gap pose 
additional challenges for deploying these methods in a live system.

\section{Proposed Method}
\vspace{-2mm}

Our proposed method focuses on processing high-resolution images efficiently, 
and is built around a fast, accurate depth densification approach based on bilateral filtering \cite{BF}. 
Bilateral filtering (BF) is a classical edge-preserving smoothing technique originally 
used for image denoising, which combines color and spatial weighting. 
A variant called joint bilateral filtering (JBF) \cite{JBF} uses a separate guidance image 
to guide the filtering. 
We leverage this concept for depth completion: 
the idea is to smooth and interpolate the sparse depth values 
in a way that respects edges in the high-resolution color image.
We implement joint bilateral filtering in a CNN-inspired manner 
to fully exploit GPU parallelism and allow the filter parameters to be trainable. 
By formulating the filtering operations as convolution-like layers, 
our method can be executed rapidly using deep learning libraries (e.g. PyTorch \cite{PyTorch}), 
and if ground-truth dense depth is available, 
parameters such as the Gaussian variances can be optimized via backpropagation. 
Importantly, our approach does not require pre-training on dense depth data.
The filter parameters can be set empirically and 
still produce good results, 
with the option to fine-tune them if data is available.

\subsection{LiDAR Data Preprocessing}
\vspace{-2mm}

Before performing depth densification, 
we preprocess the raw LiDAR point cloud data to make it more suitable for the filtering stage. 
In our system, each viewpoint’s depth image is generated 
by merging three sequential LiDAR frames from three LiDAR sensors. 
This multi-frame fusion increases point density because each LiDAR
captures slightly different parts of the scene. 
However, it also introduces two types of artifacts: 
(1) afterimages from moving objects, 
and (2) points belonging to occluded regions that the camera cannot see. 
We address these issues with the following steps.

\begin{figure}
    \centering
    \includesvg[inkscapelatex=false, width=1.0\linewidth]{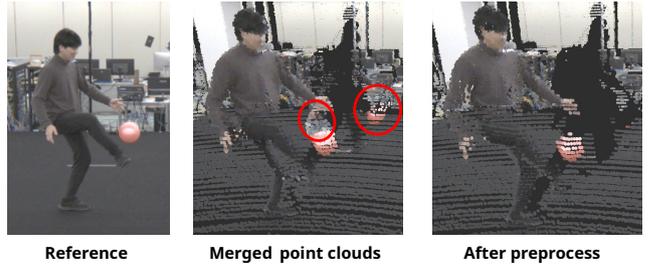}
    \vspace{-3mm}
    \caption{Example results of LiDAR data preprocess (Each point drawn with a diameter of 5 cm).}
    \label{fig:denoise}
    \vspace{-8mm}
\end{figure}

\begin{algorithm*}[bt]
    \caption{Bilateral Filtering (Depth Densification)}
    \label{alg:BF}
    \begin{algorithmic}
    \small
    \setlength{\baselineskip}{6pt}
    \Require Color image $I(x,y)$ and sparse depth map $D(x,y)$; $B(x,y)=1$ if depth present at (x,y), else 0.
    %
    \ForAll{$x, y$} \Comment{1. Compute color weight component (convolution-like operation)}
        \For{$a = -r$ to $r$ and $b = -r$ to $r$}
            \State $w(x,y,(2r+1)(r+a)+(r+b)) \gets I(x,y) - I(x+a,y+b)$
        \EndFor
    \EndFor
    \State $w \gets \exp(- \frac{w^2}{2 \sigma_c^2})$ \Comment{2. Apply color Gaussian}
    \ForAll{$x, y$} \Comment{3. Incorporate spatial weight (Hadamard product)}
        \For{$a = -r$ to $r$ and $b = -r$ to $r$}
            \State $w(x,y,(2r+1)(r+a)+(r+b)) \gets w(x,y,(2r+1)(r+a)+(r+b))*\exp(- \frac{a^2+b^2}{2 \sigma_p^2})$
        \EndFor
    \EndFor
    \State Initialize $\hat{D}, W$ to 0.
    \ForAll{$x, y$} \Comment{4. Distribute depths to neighbors (transpose convolution)}
        \For{$a = -r$ to $r$ and $b = -r$ to $r$}
            \State $\hat{D}(x+a,y+b) \gets \hat{D}(x+a,y+b) + D(x,y) w(x,y,(2r+1)(r+a)+(r+b))$
            \State $W(x+a,y+b) \gets W(x+a,y+b) + B(x,y) w(x,y,(2r+1)(r+a)+(r+b))$
        \EndFor
    \EndFor
    
    \ForAll{$x, y$} \Comment{5. Normalize the output depth}
        \State $\hat{D}(x,y) \gets \frac{\hat{D}(x,y)}{W(x,y)}$
    \EndFor
    \end{algorithmic}
\end{algorithm*}

\textbf{Removing Afterimages:}
Afterimage artifacts occur in regions where the scene content moves 
between the combined frames. 
For example, a moving person will leave behind faint “ghost” points 
from earlier LiDAR frames. 
To eliminate these, we detect moving regions via the RGB video stream. 
We compute the Euclidean distance between RGB pixel values in consecutive frames, 
and mark regions exceeding a threshold $\tau_m$ as motion areas. 
This typically highlights the contours of moving objects. 
We then apply a morphological closing operation to fill gaps in the detected motion mask, 
obtaining a cohesive region covering each moving object. 
Finally, for those regions, we discard any LiDAR points that are not from the most recent frame. 
In other words, if an old LiDAR frame’s points overlap a moving object’s current position, 
those outdated points are removed. 
This effectively erases the afterimage ghost points of moving objects.

\textbf{Removing Occluded Points:}
Due to differing LiDAR perspectives, 
the merged point cloud may include surfaces hidden from the camera. 
For example, background points captured by LiDAR might receive 
the color of a foreground object, 
causing visual artifacts. 
To fix this, we remove likely occluded points using 
depth from the merged LiDAR frames. 
An averaging filter is applied to the depth map, 
and points significantly behind the average are removed as occluded.

Fig. \ref{fig:denoise} illustrates the impact. 
The middle image shows a naive merge, 
with ghosting and color errors from motion and occlusion. 
The right image, after preprocessing, eliminates these issues. 
Ghost points and miscolored occluded surfaces are removed, 
resulting in a cleaner, more accurate point cloud.

\subsection{Depth Densification via Bilateral Filtering}
\vspace{-2mm}

Using the cleaned, merged LiDAR point cloud and the high-resolution RGB image as input, 
our method then produces a dense depth map via joint bilateral filtering. 
The bilateral filter uses the RGB image as a guide to interpolate the sparse depth values, 
ensuring that depth edges align with color edges. 
Formally, the interpolated depth at a pixel $x$, denoted $\hat{D}(x)$, is given by:

\vspace{-5mm}
\begin{equation}
    \small
    \hat{D}(x) = \frac{1}{W(x)} \sum_{x_n \in \varepsilon(x)} D(x_n) w_c(I(x_n)|I(x)) w_p(x_n|x).
    \label{eq:BF}
    \vspace{-2mm}
\end{equation}

Here, $D(x_n)$ is the observed depth at a neighboring pixel $x_n$, 
and the sum is taken over the set $\varepsilon(x)$ of neighboring pixels 
around $x$ that have depth observations. 
The weights $w_c$ and $w_p$ measure the similarity between pixel $x$ and each neighbor $x_n$ in terms of color and spatial distance, 
respectively, and $W(x)$ is a normalization factor. 
The weighting functions are defined as Gaussian kernels:

\vspace{-6mm}
\begin{equation}
    w_c(I(x_n)|I(x)) = \exp(-\frac{||I(x_n) - I(x)||^2}{2 \sigma_c^2}),
    \vspace{-2mm}
\end{equation}
\begin{equation}
    w_p(x_n|x) = \exp(-\frac{||x_n - x||^2}{2 \sigma_p^2}).
    \vspace{-1mm}
\end{equation}

Here $I(x)$ and $I(x_n)$ are the RGB color vectors at pixels $x$ and $x_n$, 
and $\sigma_c$, $\sigma_p$ are parameters controlling the fall-off of weight 
with color difference and spatial distance, respectively. 
The normalization term $W(x)$ is then:

\vspace{-4mm}
\begin{equation}
    W(x) = \sum_{x_n \in \varepsilon(x)} w_c(I(x_n)|I(x)) w_p(x_n|x).
    \vspace{-2mm}
\end{equation}

Given these definitions, 
Eq. \ref{eq:BF} computes a weighted average of the observed depths in the neighborhood of $x$, 
where closer neighbors in color  and 
space contribute more, yielding the filled-in depth $\hat{D}(x)$.

We implement this filtering process using operations 
that can be mapped to efficient GPU kernels. 
Algorithm \ref{alg:BF} outlines the computation steps for the bilateral filtering. 
In essence, the algorithm performs three main phases: 
a convolution-like aggregation of color differences, 
an element-wise (Hadamard) multiplication to incorporate spatial weighting, 
and a transpose-convolution step to distribute depth contributions to neighboring pixels, 
followed by normalization.

In Algorithm \ref{alg:BF}, 
$w[x,y,\cdot]$ represents the set of weights for all neighbor offsets around pixel $(x,y)$. 
Step 1 computes the intensity difference between the center pixel and each neighboring pixel 
within a radius $r$. 
Step 2 applies the color Gaussian $w_c$ to these differences. 
Step 3 multiplies by the spatial Gaussian $w_p$ for each neighbor offset $(a,b)$. 
In step 4, 
each observed depth value $D(x,y)$ is spread to its neighboring positions $(x+a, y+b)$, 
accumulated into $\hat{D}$ 
with the precomputed weight $w[x,y,idx]$. 
Simultaneously, a binary mask $B(x,y)$ indicating whether a depth value exists at $(x,y)$ 
is spread in the same way to accumulate the total weight $W(x+a,y+b)$ 
for normalization. 
In step 5, the accumulated depths are normalized by these weights to produce the final dense depth map $\hat{D}$.

This implementation shows that bilateral filtering can be 
achieved by a combination of standard operations 
(convolution, element-wise multiplication, and transpose convolution) 
commonly found in CNN frameworks. 

Our bilateral filter-based depth completion can be implemented very efficiently on a GPU. 
We have written it using PyTorch \cite{PyTorch}, 
taking advantage of optimized tensor operations. 
As a result, the entire filtering can run in real-time even on high-resolution images. 

\section{Experimental Results}
\vspace{-2mm}

We integrated the proposed depth densification method 
into our real-time spatial transmission system 
and evaluated its performance qualitatively and in terms of speed. 
The test scene consisted of a person moving in an instrumented area surrounded 
by cameras and LiDARs. 
The RGB cameras (1920×1080) captured at 30 fps, 
and the LiDAR point clouds were updated at an effective 30 fps 
by merging three 10 fps LiDAR streams, 
yielding about 100,000 points per frame. 
We deployed seven viewpoints around the subject, 
each viewpoint consisting of one camera and three LiDAR units 
to cover the scene from multiple angles. 
Each viewpoint processed its data in parallel, 
performing the depth densification locally, 
and then the resulting dense colored point clouds 
from all views were combined into a single 3D point cloud representing the scene.

\begin{figure}[t]
    \centering
    \includesvg[inkscapelatex=false, width=0.8\linewidth]{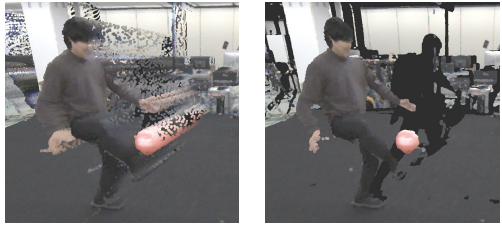}
    \caption{Contour removal process (Each point drawn with a diameter of 5 cm).}
    \label{fig:contour}
    \vspace{-8mm}
\end{figure}

After the densification process, each view tends to produce significant noise along object contours like Fig.\ref{fig:contour}.
To suppress these artifacts, we remove unreliable depth values near the contours by applying a simple thresholding method based on the gradient of the depth map.
This denoising procedure is applied to both our proposed method and BP-Net for fair comparison.

In our proposed method, we adopt a 2-stage filtering approach to achieve efficient and accurate depth completion.
In the first stage, filtering is performed on a downsampled depth map, reduced to 1/3 of the original resolution in both vertical and horizontal dimensions using pooling.
In the second stage, filtering is applied to the depth map after restoring it to the original resolution.
We use a filter radius of 4 in the first stage and a radius of 2 in the second stage.

\begin{figure}[t]
    \centering
    \includesvg[inkscapelatex=false, width=1.0\linewidth]{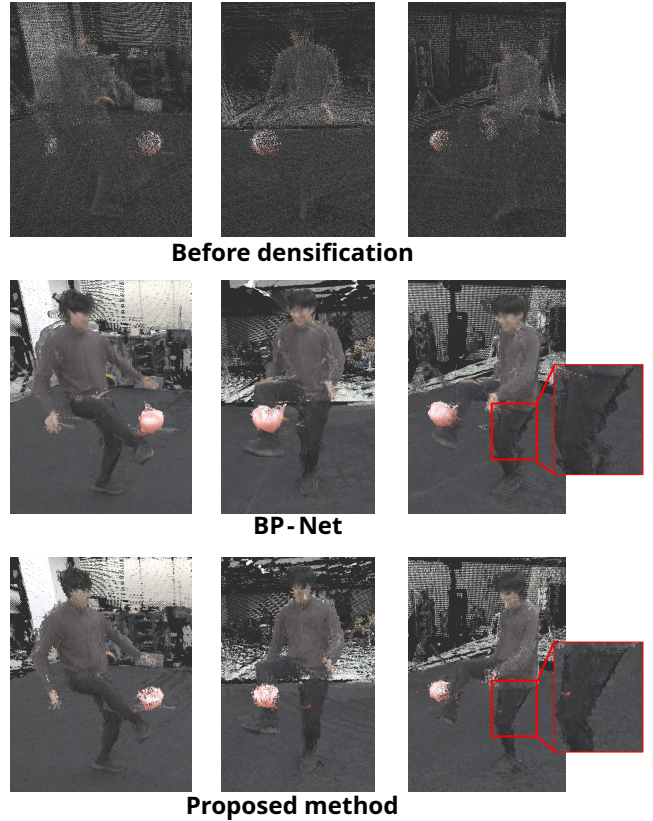}
    \vspace{-4mm}
    \caption{Example results of real-time densification (Each point drawn with a diameter of 1 cm).}
    \label{fig:result}
    \vspace{-8mm}
\end{figure}

Fig. \ref{fig:result} shows an example of the obtained results. 
The figure compares the point cloud before densification,
the after densification result using our method, 
and the result produced by BP-Net \cite{BPNet}. 
For the BP-Net result, 
we used a publicly available implementation 
with a model pre-trained on the NYU Depth v2 indoor dataset \cite{NYUDepthV2}. 
From the figure, it is clear that our method successfully 
produces a much denser 3D point cloud of the person and the environment. 
The reconstructed surfaces using our approach appear complete 
and well-aligned across the multiple viewpoints. 
In contrast, BP-Net does increase the density of the point cloud 
to some extent and the depth maps from each view appear smooth; 
however, we observed that BP-Net tended 
to introduce a slight depth offset for the entire subject. 
When the depth outputs from different cameras are merged, 
these per-view depth misalignments cause the geometry 
to appear as ghosted double surfaces 
in the combined point cloud. 
Our method, despite its simplicity, yields depth maps 
that are properly aligned across all viewpoints, 
resulting in a clean integrated 3D reconstruction. 
The visual quality of our dense point cloud is on par with, 
or better than, the learning-based BP-Net in this multi-view scenario, 
while using a much simpler filtering model.

Table \ref{tab:time} summarizes the processing speed of our method compared to BP-Net. 
We measured the per-frame computation time on a Linux PC equipped with 
an NVIDIA GeForce RTX 4090 GPU, for Full HD (1080p) input images. 
As shown, BP-Net took around 420 ms per frame, 
whereas our method took only about 23 ms. 
In practice, our implementation easily achieves 
30 fps real-time performance on 1080p data, 
even with the processing for seven viewpoints running in parallel.
The speed advantage of our approach demonstrates 
its suitability for live systems, where low latency is critical.

\begin{table}[htb]
    \centering
    \vspace{-3mm}
    \caption{Processing Time (per frame)}
\begin{tabular}{|l|l|}
      Method & Processing Time \\
      \hline
      BP-Net \cite{BPNet} (320x240) & 20 ms (reported) \\
      BP-Net (1080p) & 420 ms \\
      Proposed Method & \textbf{23 ms}
\end{tabular}
    \vspace{-2mm}
\label{tab:time}
\end{table}


\section{Conclusion}
\vspace{-2mm}

We presented a method for real-time 3D shape acquisition 
in a spatial transmission system 
that reproduces a remote environment by combining camera and LiDAR data. 
The proposed approach is a relatively simple, 
filter-based method built upon bilateral filtering, 
yet it achieves high-quality, high-density 3D point clouds at real-time speeds. 
We demonstrated that this method can produce visually pleasing results 
and significantly outperforms more complex state-of-the-art depth completion techniques 
in processing speed, 
while still maintaining competitive output quality. 
These properties make our approach well-suited 
for live 3D telepresence systems, 
where dense 3D capture and low latency are both essential.

\bibliographystyle{unsrt}
\bibliography{ref}

\begin{thebibliography}{10}

\bibitem{BPNet}
Jie Tang et~al.
\newblock {Bilateral Propagation Network for Depth Completion}.
\newblock In {\em Proc. of CVPR}, pages 9763--9772, 2024.

\bibitem{holoportation}
Sergio Orts-Escolano et~al.
\newblock {Holoportation: Virtual 3D Teleportation in Real-time}.
\newblock In {\em Proc. of the 29th annual symposium on user interface software and technology}, pages 741--754, 2016.

\bibitem{dou2016fusion4d}
Mingsong Dou and othres.
\newblock {Fusion4d: Real-time performance capture of challenging scenes}.
\newblock {\em ACM Transactions on Graphics (ToG)}, 35(4):1--13, 2016.

\bibitem{volumetric}
Oliver Schreer et~al.
\newblock {Capture and 3D Video Processing of Volumetric Video}.
\newblock In {\em Proc. of ICIP}, pages 4310--4314, 2019.

\bibitem{DepthCompSurvey2022}
Junjie Hu et~al.
\newblock {Deep depth completion from extremely sparse data: A survey}.
\newblock {\em IEEE Trans. on Pattern Analysis and Machine Intelligence}, 45(7):8244--8264, 2022.

\bibitem{RGBGuidedSurvey2024}
Xin Qiao et~al.
\newblock {Rgb guided tof imaging system: a survey of deep learning-based methods}.
\newblock {\em International Journal of Computer Vision}, 132(11):4954--4991, 2024.

\bibitem{DAGF}
Zhong Zhiwei et~al.
\newblock {Deep Attentional Guided Image Filtering}.
\newblock {\em IEEE Trans. on Neural Networks and Learning Systems}, 2023.

\bibitem{bai2020depthnet}
Lin Bai et~al.
\newblock {DepthNet: Real-time LiDAR point cloud depth completion for autonomous vehicles}.
\newblock {\em IEEE Access}, 8:227825--227833, 2020.

\bibitem{long2024sparsedc}
Chen Long et~al.
\newblock {SparseDC: depth completion from sparse and non-uniform inputs}.
\newblock {\em Information Fusion}, 110:102470, 2024.

\bibitem{NYUDepthV2}
Nathan Silberman et~al.
\newblock {Indoor segmentation and support inference from rgbd images}.
\newblock In {\em Proc. of ECCV}, pages 746--760. Springer, 2012.

\bibitem{BF}
Carlo Tomasi and Roberto Manduchi.
\newblock {Bilateral Filtering for Gray and Color Images}.
\newblock In {\em Proc. of ICCV}, pages 839--846. IEEE, 1998.

\bibitem{JBF}
Chunxia Xiao and Jiajia Gan.
\newblock {Fast Image Dehazing using Guided Joint Bilateral Filter}.
\newblock {\em The Visual Computer}, 28:713--721, 2012.

\bibitem{PyTorch}
{PyTorch Foundation}.
\newblock {PyTorch}.
\newblock \url{https://pytorch.org/}.

\end{thebibliography}

\end{document}